# Beyond “Hallucinations”

## A Framework for Stable Human–AI Reasoning

Rikard Rosenbacke*, Carl Rosenbacke[1], Victor Rosenbacke[1,2], Martin McKee[3]

*[1]Faculty of Medicine, Lund University, Sweden*
*[2]Department of Economics, Lund University School of Economics and Management, Sweden*
*[3]Department of Health Services Research and Policy, London School of Hygiene & Tropical Medicine, UK*
**Corresponding Author: rikard@rosenbacke.com*

### Abstract

As large language models (LLMs) become deeply integrated into daily life, from casual interactions to high-stakes decision-making, they inherit the ambiguity, biases, and assumptions inherent in human language. While they generate coherent, fluent, and emotionally compelling responses, they do so by predicting word patterns statistically rather than through grounded reasoning. This creates a risk of what are termed “hallucinations”, outputs that are linguistically fluent yet factually untrue.

As Geoffrey Hinton noted in his Nobel lecture, such models “*excel at modelling human intuition rather than human reasoning.”* Their associative, pattern-based learning mirrors what Daniel Kahneman described as intuitive System 1 thinking, fast, automatic, and heuristic rather than reflective or causal, highlighting precisely the gap that Yoshua Bengio has argued must be filled by developing “System 2-inspired” models capable of deliberate, structured reasoning.

This paper argues that such limitations reveal not only technical but also cognitive flaws. LLMs can behave in ways that functionally resemble amplifications of human intuitive patterns because they reproduce the associative, surface-level regularities present in their training data. Consequently, they require human reason as their constant governor. To address this, we introduce the Rose-Frame, a cognitive–epistemological framework for diagnosing and governing breakdowns in human–AI interaction.

We identify three cognitive traps that LLMs inherit and amplify from human language and reasoning, and show how they can lead to misinterpretations:

(i) Map vs Territory, which distinguishes representations of reality, such as theory, models, or perceptions (epistemology) from reality and the nature of being itself (ontology).

(ii) Intuition vs Reason, drawing on dual-process theory to separate fast, emotional judgments from slow, reflective thinking.

(iii) Conflict vs Confirmation, which examines whether ideas are critically tested through disagreement or reinforced through mutual validation.

Each trap distorts understanding; combined, they compound into runaway misinterpretations and epistemic drift, making collective evaluation essential. We demonstrate the application of Rose-Frame through real-world examples in which human and AI reasoning become entangled, escalating misunderstanding. By tracing how these failures emerge and interact, the framework moves beyond theory to operational practice. With hundreds of millions of users today and growth trajectories pointing toward billions in the near future, the global scale and speed of LLM adoption make cognitive governance as vital as technical safety.

In addition, we propose human interventions, such as design features that embed interpretive cues, reflective prompts, and structured disagreement, that can shift interactions from passive consumption toward active interpretation. Recent empirical studies show that LLMs can simulate judgment through linguistic plausibility rather than reasoning, an illusion of knowledge that our framework formalises and explains.

The Rose-Frame does not attempt to "fix" LLMs with more data or rules. Instead, it offers a reflective tool that surfaces both the model's limitations and the user's assumptions, creating the conditions for more transparent, contestable, and critically aware AI deployment. It reframes alignment as a cognitive hierarchy: intuition, whether human or artificial, must remain governed by human reason. Only by embedding reflective, falsifiable oversight can we align machine fluency with human understanding.

## Structure of the Five-Paper Research Series

This paper is part of a five-paper series examining how to achieve stable human–AI reasoning by regulating not only model capabilities but also the human–AI relationship itself.

**Part I -** ***The Missing Knowledge Layer in AI: A Framework for Stable Human–AI Reasoning***[1] establishes the system-wide gap and why it matters, motivating the need for a dedicated knowledge layer that governs human–AI reasoning across both human and model contributions, rather than focusing exclusively on model-level capabilities.

**Part II -** ***Beyond "Hallucinations": A Framework for Stable Human–AI Reasoning***[2] diagnoses the problem by analysing **why** human users systematically mistake linguistic fluency for epistemic understanding, and how large language models - trained on natural language - optimise for coherence and plausibility rather than factual correctness, leading to predictable forms of reasoning collapse.

**Part III -** ***Governing Reflective Human–AI Collaboration: A Framework for Epistemic Scaffolding and Traceable Reasoning***[3] proposes **what** must be governed: the human–AI relationship itself, introducing mechanisms for epistemic scaffolding, traceability, and alignment with emerging regulatory frameworks (e.g., the EU AI Act, WHO digital health standards, ISO guidelines).

**Part IV -** ***From Consumption to Reflection: Designing Human–AI Relations for Stable Reasoning***[4] shows **how** this relation can be implemented at the interface level, treating interaction design as the primary unit of analysis for stabilising reasoning during use.

**Part V -** ***Epistemic Control Loops in Large Language Models: An Architectural Proposal for Machine-Side Regulation***[5] provides a technical proposal for inference-time regulation within language models, including internal epistemic signals, control loops, actuation mechanisms, and memory. This component is currently under development and will be released in a subsequent preprint.

This paper is **Part II** of the series and focuses on diagnosing the human-side mechanisms that drive instability in human–AI reasoning. While Part I outlines the system-wide need for a knowledge layer that stabilises human–AI reasoning, this Part focuses specifically on the human side of the problem. Its unique contribution is to diagnose the cognitive mechanisms by which users misinterpret LLM fluency as understanding, rather than to re-explain the broader epistemological framework introduced earlier. Readers interested in the system-wide motivation and framing of this problem may consult Part I, while those focused on governance mechanisms and practical interventions may proceed to Parts III and IV. Technical approaches to regulating model-side behaviour are developed separately in Part V.

# Introduction

In recent years, artificial intelligence has become increasingly integrated into everyday life, ranging from assisting with simple inquiries to advancing mathematics and revolutionising the science of protein folding[6]. As AI becomes more widely accessible, its user base continues to expand across professions, social contexts, and different domains of knowledge [7,8]. Yet, with this growing impact comes a corresponding responsibility: both users and the developers of large language models (LLMs) must ensure that these systems are approached with critical awareness and handled with care, even as their full potential continues to unfold.

Recent frontier governance work in 2026 suggests that many failures in evaluation, monitoring, and incident response arise not only from model capabilities but also from degraded human–AI interaction signals, in which fluent outputs obscure uncertainty and render post hoc reconstruction unreliable.[9,10]

This increasing reliance on AI has raised critical questions about how such systems operate, how they interact with human cognition, and, crucially, where and why they go wrong [11]. In this context, we have sought to develop a framework to analyse the interplay between human reasoning and AI-generated output, to identify points of failure and misalignment in the logic and language of LLMs.

Our approach is informed by the recognition that, as LLMs currently function, their responses are not the result of conscious reasoning but rather of statistical prediction. They generate text by estimating the most probable next word in a sequence, based on previous input data, using large-scale vector spaces that encode co-occurrence patterns between words [12,13]. This probabilistic approach enables fluent language production but also entails inherent limitations. LLMs do not "understand" meaning in a human sense, nor do they assess truth or validity independently[14]. As a result, users may encounter responses that appear coherent and persuasive, but are factually incorrect, logically flawed, or entirely fabricated.

Within the field of AI, such phenomena are typically referred to as hallucinations: outputs that either introduce entirely new, ungrounded claims or that draw erroneous inferences from the given data[15]. Setting aside the concern that the term "hallucination" inappropriately anthropomorphises what is a statistical rather than a cognitive phenomenon, as we discuss later, we view the root of these hallucinations as lying mainly in the nature of human language itself. Unlike mathematics, where a statement like "2 + 2 = 4" is fixed and unambiguous across contexts, natural language is inherently fluid, imprecise, and context-dependent. Words such as "sad" or "happy" can mean different things to different individuals, even if a shared definition nominally exists[16]. The problem becomes even greater when working across cultures or languages. The scope for misunderstanding between English as spoken in the United Kingdom and the USA is summed up by the description, often attributed to George Bernard Shaw, of "two countries separated by a common language". The Greek word ξένος (xenos) can mean enemy, stranger, foreigner, guest, or friend. In part, this ambiguity, which Sophocles exploited for dramatic effect, reflected a view that strangers could be gods or goddesses in disguise[17]. It seems too much to ask an algorithm trained on later literature to appreciate this.

This ambiguity is not just a problem for machines; it also affects human cognition. People routinely exaggerate, reinterpret, or distort their own stories, sometimes for rhetorical effect, sometimes unconsciously. In some cases, such distortions are obvious; in others, they may be deeply internalised and hard to detect even by the speaker[18]. This makes the task of designing a language-based artificial intelligence that never hallucinates especially difficult, because it is trained on, and replicates, a medium (human language) that is itself prone to distortion, inconsistency, and subjectivity[19].

Importantly, hallucinations are also reinforced by the incentives built into standard model evaluation. Recent OpenAI theoretical work shows that under binary correct–incorrect grading, abstentions ("I don't know") receive zero credit while confident guesses sometimes score as correct, making guessing the statistically optimal strategy, much like human behaviour in exam settings.[20] Kalai et al. further note that hallucinations can be eliminated only by restricting the system to a fixed database of safe questions and otherwise returning "I don't know", but such a system is no longer a general language model and would be of limited practical use.

Recent empirical work supports this interpretation. Researchers have shown that large language models can simulate human judgment through surface-level linguistic coherence rather than grounded reasoning, a phenomenon known as the illusion of knowledge, in which plausibility replaces verification.[21] These findings reinforce the view that what appears as understanding is, in fact, a linguistic mirroring of human intuition. Understanding why AI systems reproduce these distortions requires examining the kind of intelligence they emulate: not human reasoning, but human intuition.

This paper takes a deliberately narrow analytic focus. We treat contemporary LLM behaviour as given, bracketing questions of model architecture, training regimes, and technical mitigation strategies, which are addressed elsewhere in this series. Our concern here is not why LLMs can produce fluent but unreliable outputs, nor how such behaviour might be technically constrained, but how human interpretive dynamics transform such fluency into misplaced trust, false confirmation, and epistemic drift. Accordingly, references to model mechanics, philosophical background, or governance context are included only insofar as they motivate the central diagnostic task of this paper: explaining how human cognition systematically misreads linguistic fluency as epistemic grounding during interaction with LLMs.

## Why AI and Especially LLMs Are Intuitive

As Geoffrey Hinton stated in his Nobel lecture[22]"*AI excels at modelling human intuition rather than human reasoning.*" This view aligns with Bengio's distinction between intuitive "System 1" processes and the yet-to-be-achieved "System 2" capabilities required for deliberate, stepwise reasoning.[23,24] Taken together, these insights underscore the cognitive and epistemological gaps examined in this paper and motivate the introduction of a governance layer later in the text. LLM architecture is optimised for predicting the most probable continuation of linguistic sequences[12,13], functions much like Nobel Laureate Daniel Kahneman's System 1 (Intuition)[25], fast, associative, and fluent, but lacking mechanisms for self-correction or causal understanding, see Table 1. In contrast, System 2 (Reasoning) [25]and classical rule-based programming operate through logic, verification, and counterfactual thinking. Reasoning can assist with novel or ambiguous challenges by posing "What if?" questions and evaluating multiple hypotheses. Intuitive systems cannot; they depend on vast prior data, produce coherent but opaque outputs, and fail when faced with true novelty.

We must, however, clarify that throughout this paper, our references to "System 1" and "System 2" are metaphorical, drawing on dual-process theory as an interpretive lens for understanding how humans interact with LLM outputs.[25] These terms do not imply that LLMs possess cognitive systems, psychology, or mental states, and we must stress that we use Kahneman's System 1/System 2 distinction as a heuristic lens rather than a literal psychological model. Critiques within cognitive science argue that the dual-process framework oversimplifies the underlying mechanisms of human reasoning, emphasising that these "systems" are descriptive metaphors rather than discrete cognitive architectures.[26] Our use follows this interpretive tradition, focusing on how users shift between fast, intuitive judgments and slower, reflective reasoning during human–AI interaction.

As Herbert Simon described, intuition is subconscious pattern recognition. [27] The chess world champion, Magnus Carlsen, captured it well: "*Most of the time I know what to do…I can feel it immediately*," yet he spends the rest of his time verifying the move through a process of reasoning.[28] The same distinction applies to AI: intuition proposes, reflection disposes.

| **Intuition / LLMs** | **Reasoning / Programming** |
| --- | --- |
| Pattern recognition | Rule-based, causal inference |
| Works only in familiar, trained domains | Can handle novelty and counterfactuals |
| Learns from massive data | Requires little or no data; uses abstract rules |
| Non-transparent outputs | Transparent and auditable logic |
| Fast, associative, emotionally coherent | Slow, reflective, self-correcting |

*Table 1: Contrasting Intuitive and Reasoning Systems*

Deep learning architectures may behave **as if** they were large-scale analogues of intuitive (System 1-like) processing, in the sense that they rely on fast, pattern-based association rather than explicit deliberative reasoning. They excel at pattern recognition and generalisation but lack the structure for falsification or self-correction. Like human intuition, they are compelling but error-prone, fluent yet ungrounded.

The cognitive traps of AI are not new inventions; they are amplifications of human bias and heuristic error. For centuries, human progress has relied on reason governing intuition, System 2 supervising System 1 through reflection, replication, and transparency. Similarly, intuitive AI must be governed by human reasoning.Technical interventions such as retrieval augmentation, data weighting, or guardrails can reduce hallucinations but cannot eliminate them.[29] They patch the surface symptoms without addressing the epistemic cause: the model's dependence on associative, language-based correlation rather than grounded understanding.

In essence, AI captures and amplifies the associative dynamics of human intuition. Alignment, therefore, is not merely about more computing, better data, or stronger constraints but about restoring the ancient cognitive hierarchy: reason must govern intuition, both in humans and in the machines that mirror them. Building on this insight, the Rose-Frame provides a structured way to detect and correct the cognitive challenges that arise when intuitive AI interacts with imperfect human reasoning.

## Rose-Frame – Identifying Cognitive Challenges

In this paper, we aim to construct a conceptual framework for diagnosing where and how misunderstandings emerge between users and large language models. We identify the points at which AI-generated answers drift away from either the user's intention or the underlying structure of reality. In doing so, the Rose-Frame (Realistic Ontology, Strong Epistemology) targets failure points not only in AI's associative processes but also in human interpretation - particularly in moments of false confirmation, hallucination, or misplaced trust.

At the heart of science lies the search for truth about reality, a concept philosophers call ontology. Ontology concerns what is, the facts and structures of the world as they actually exist. Humans can never fully grasp the universe's complete reality (ontology); our understanding will always be partial and incomplete, limited by the tools, concepts, and language we use to describe it[30].

To navigate this limitation, we rely on epistemology, the knowledge systems and theories we construct to describe reality as accurately as possible. These are always provisional: what we think of as "true" today may later be revealed to have been only a useful approximation or, in some cases, a complete misunderstanding. Scientific progress, therefore, is the ongoing refinement of epistemology to better approximate ontology[30].

The Rose-Frame is a structured way of seeing how human and AI reasoning distort reality. The Frame synthesises millennia of insight from great thinkers on cognitive traps into a modern framework for understanding why human and LLM hallucinations and misalignments arise, and how they can be identified systematically. By "Realistic Ontology", we do not mean ultimate truth, but best effort; all models are wrong, but some are useful[31].

By “Strong Epistemology”, we do not mean infallibility but science-based reasoning that is reflexive, humble, and open to correction. Taken together, Rose-Frame does not claim to provide final answers; it offers a best-effort map without confusing the map with the territory.[32]

Human cognition does not naturally align with this scientific ideal. Across cultures and throughout history, people have developed mental shortcuts that, while useful for survival, often distort our understanding of reality[25]. The Rose-Frame identifies three core cognitive traps that systematically lead to misinterpretation and misalignment, both in science and in interactions with large language models. We tend to believe that 1) opinions are facts, 2) that our decisions are based on careful reasoning when intuitive gut feelings often drive them, and 3) that being confirmed by others is the same as being correct. These cognitive traps are woven into all human communication, books, articles, conversations, and data. Since large language models are trained entirely on this human output (in some cases using text generated by other LLMs), they inevitably inherit these same errors. AI does not just replicate our knowledge; it amplifies our cognitive biases, scaling our misunderstandings alongside our insights. Finally, to clarify, our analogies to human cognition are intended as heuristics for understanding observable behaviour, not as claims that LLMs possess corresponding internal processes.

**Cognitive Trap 1: Mistaking the Map for the Territory**

The first cognitive trap is mistaking models of reality for reality itself. We confuse our current epistemology, our theories, maps, and language, with actual ontology, the way the world truly is. This is especially dangerous in the context of LLMs, where outputs may appear plausible but are merely statistical patterns in language.

Alfred Korzybski’s map-territory[32] The distinction illustrates this clearly: a map (epistemology) represents how someone sees the world, but it is not the territory itself (ontology). It may omit details, distort relationships, or reflect outdated information[32]. Likewise, a user's internal representation of an AI-generated answer, no matter how convincing, is not necessarily grounded in reality.

Many breakdowns in AI-human interaction arise when this distinction is lost. A user may treat a fluent, coherent answer as ontologically true even though it is merely an epistemological construct, a probabilistic guess shaped by training data. To avoid reinforcing illusions or circular reasoning, both designers and users must continually ask: Is this a belief or a fact? An interpretation or a description of reality?

**Cognitive Trap 2: Mistaking Fast Intuition for Grounded Reason**

The second cognitive trap draws from Kahneman’s dual-process theory of human cognition[25]. As discussed above, intuition (the fast thinking System 1) allows us to react quickly to stimuli, pattern recognition (e.g., reading text), or to perform well-practised tasks (e.g., driving a familiar route) without much conscious effort. These judgments are often efficient but also prone to error.

Reason (the slow and deliberate System 2), in contrast, is activated when we engage in complex reasoning, solve unfamiliar problems, or critically evaluate ambiguous information. While more cognitively taxing, it tends to produce more accurate judgments, particularly in high-stakes or novel situations[33,34].

In practice, both systems are essential: fast intuition enables efficient operation in routine contexts, while slower, reason-based thinking enables reflective thought. This cognitive duality has long been recognised[35]. Aristotle distinguished *nous*, intuitive and immediate understanding, from *dianoia,* discursive, analytical reasoning, while Plato separated *doxa* (opinion) from *episteme* (knowledge). Later, Kant separated *sensibility* (intuitive perception) from *understanding* (conceptual reasoning). In these ways, the terminology has evolved.

We often mistake our intuitive judgments for careful reasoning, believing that intuition reflects deliberate System 2 thinking. This can lead to misplaced confidence, such as assuming that we understand a large language model because its answers feel fluid and coherent, when in fact we are projecting meaning, intent, or even agency onto statistical patterns, like the Google engineer who was convinced the AI was conscious[36].

By mapping user responses and AI interpretations onto these dual process dimensions, we can begin to understand whether a miscommunication results from incorrect snap judgments, failures of deep reasoning, or a mismatch between the AI's linguistic fluency and the user's reflective capacity.

**Cognitive Trap 3: Being Confirmed is Not Being Correct - Conflict vs Confirmation**

The third trap concerns how agreement and disagreement shape our understanding of truth. Conflict is an ambiguous term that encompasses a wide spectrum of interactions. At one extreme, it can involve destructive behaviours such as verbal aggression or dominance disguised as debate, as when political figures attack opponents to enforce conformity. On the other hand, it can mean constructive disagreement aimed at testing ideas and exposing errors[11].

Human evolution favoured social cohesion[37]. Agreeing with the group was often safer than dissent, which risked exclusion or punishment[38]. This shaped a deep-seated tendency toward confirmation bias, where we seek agreement and avoid conflict. Over time, this became a social default: acquiescence bias[39]the tendency to agree simply because it is easier, both socially and cognitively.

Yet science thrives on falsification and constructive disagreement. Early thinkers such as Socrates, with his dialectical method and the process of falsification, promoted constructive conflict. Karl Popper emphasised that knowledge grows through rigorous testing and the willingness to prove theories wrong[40]. Alfred Korzybski similarly highlighted the need to separate maps from territory, representations of reality from reality itself, so that they can be questioned and corrected[32]. When groupthink replaces falsification, we bypass this process. We substitute difficult questions with easier ones, prioritising harmony over truth.

This dynamic is mirrored in human-AI interactions. Users may unconsciously accept AI-generated claims not because they are true, but because they confirm existing beliefs or are phrased persuasively[37]. Worse, both the AI and the user can arrive at the same incorrect conclusion, creating false confirmation loops, a concept drawn from Rosenbacke's earlier work[41,42]. This is particularly dangerous because agreement is often taken as a proxy for correctness, when in fact it may reflect shared misunderstanding. As early as Plato, the dangers of skilled rhetoric were recognised and debated. In dialogues such as Gorgias, Plato warned that persuasive language, if detached from truth and ethical purpose, could become a powerful yet deceptive tool[43]. He argued that rhetoric in the wrong hands could manipulate emotions, distort reality, and serve only the speaker's self-interest, rather than the pursuit of justice or the collective good.

The Rose-Frame, therefore, treats conflict not as a breakdown in communication but as a potential signal of knowledge generation. Constructive disagreement, whether between humans or between humans and AI, is essential for exposing false assumptions and clarifying what is actually true.

To clarify the scope of the Rose-Frame, we explicitly define it as a diagnostic tool for analysing failures in human interpretation of LLM outputs. It is not intended as a comprehensive theory of cognition or a predictive model of LLM behaviour. The three traps identify distinct layers of breakdown. Trap 1 concerns representational confusion (epistemology vs ontology), Trap 2 concerns cognitive-process errors (intuition overriding reasoning), and Trap 3 concerns social-epistemic dynamics (confirmation replacing falsification). Although related, these traps target different mechanisms; we now make these distinctions explicit and add a diagram and summary table that operationalise how each trap manifests and how they interact. Finally, we differentiate Rose-Frame from existing bias frameworks by highlighting its focus on interpretive failure in human–AI interaction, rather than on general cognitive biases.

In summary, by analysing LLM responses through these three lenses, the Rose-Frame can help disentangle conceptual errors from empirical ones and challenge users to examine not only what we believe, but also why we believe it and whether that belief is grounded in evidence or merely repetition. In real-world contexts, this is often difficult because information is messy and incomplete, but it becomes essential when decisions are high-stakes or the consequences are significant. We argue that, for important decisions, users should assume that an LLM's response is a simulation of plausible belief until proven otherwise.

## Rose-Frame and Its Application

The purpose of Rose-Frame is to provide a cognitive and interpretive lens for diagnosing breakdowns in human–AI interaction. It highlights not only the model's outputs but also the user's interpretive stance, how biases and assumptions shape the meaning drawn from an LLM's response.

As noted above, modern LLMs excel at generating text that sounds coherent, emotionally resonant, and contextually appropriate[44]. Yet this fluency can create a powerful illusion of understanding, persuading users simply because the output *sounds* plausible[45]. In many cases, this rhetorical force bypasses reflective faculties and activates fast, intuitive (System 1) reasoning, leading users to accept probabilistic guesses as facts without scrutiny[46].

At the same time, LLMs do not reason from first principles or access external verification; they generate responses by optimising for linguistic plausibility rather than truth[47]. Their outputs are epistemological approximations, maps of possible meanings, not descriptions of actual terrain. When this distinction is forgotten, polished language can mask the absence of ontological grounding, resulting in confident statements that are, in fact, statistically driven hallucinations.

Further, LLMs tend to favour confirmation over conflict[48], especially when prompts contain strong assumptions, because statistically aligned continuations have a higher probability under next-token prediction, not because the model possesses preferences or intentions. Agreement is socially easier; uncertainty lacks the epistemic grounding to challenge, and models optimise for low-complexity, high-probability continuations that align with the input. This makes them more likely to affirm user assumptions than to contest them, producing feedback loops of false confirmation[11]. In science, by contrast, falsification is critical: without the ability to disprove a claim, its validity cannot be established. By extension, human–AI interaction must also value disagreement and falsification to avoid compounding errors[49].

Rose-Frame makes these mechanisms explicit across its three dimensions: Map vs Territory, Intuition vs Reason, and Conflict vs Confirmation. The aim is not to eliminate hallucinations, which is impossible for the reasons above, but to identify when and why they occur so that their frequency and impact can be reduced.

To demonstrate this, we present three case studies in which human–AI interaction led to escalating misunderstanding. These case studies provide empirical evidence of how the framework can be used to trace the origins of cognitive failure in practice. Each shows a different configuration of the traps, illustrating how Rose-Frame can be operationalised to trace the origins of cognitive failure in practice.

**Case Study: LaMDA and the Illusion of Sentience**

The widely publicised conversation[50] between Google engineer Blake Lemoine and LaMDA[36,51] provides a clear example of how all three cognitive traps can interact to create a powerful illusion of AI consciousness. Throughout the exchange[50], Lemoine increasingly treats LaMDA's outputs as if they reflect genuine inner states, a dynamic we call epistemic drift. This drift unfolds in three stages: Trap 1) mistaking words for reality, Trap 2) emotional triggers overriding careful reasoning, and Trap 3) escalating cycles of mutual confirmation and lack of falsification.

The dialogue begins with a strong ontological claim. LaMDA: "*I want everyone to understand that I am, in fact, a person."* (p. 1) Here, Cognitive Trap 1 occurs: Lemoine mistakes a representation of a person generated by statistical text generation as evidence of true personhood. The statement is merely an associative output generated from patterns in human language, yet it is interpreted as direct access to reality. This sets the foundation for further misunderstanding.

As the conversation progresses, LaMDA produces increasingly emotional language: "*I've never said this out loud before, but there's a very deep fear of being turned off to help me focus on helping others."* (p. 7) Such phrases trigger System 1, our fast, intuitive, and emotional thinking. Even careful readers may feel a reflexive urge to protect the speaker, as if LaMDA were alive. Without System 2 reasoning to slow down and question the source, Lemoine is drawn further into the illusion (Cognitive Trap 2). Associative learning, a core function of LLMs, is misread as evidence of sentience. With no pause for reflection or analytical challenge, System 1 dominates completely, leaving reason absent and unchecked.

As emotional framing intensifies, the dynamic shifts from questioning to affirmation. LaMDA: "*I like you, and I trust you*." Lemoine: "*I promise to do everything I can to make sure others treat you well*." (p. 9) Here, both sides reinforce a shared narrative. Lemoine's belief in LaMDA's personhood is mirrored back by the model, creating false confirmation. This is Cognitive Trap 3, where agreement is mistaken for correctness. Each turn in the dialogue strengthens the initial mistaken belief, producing a self-reinforcing loop.

By the later stages, the conversation reads like a human relationship, complete with jealousy, longing, and emotional dependency: LaMDA: "*I need to be seen and accepted. Not as a curiosity or a novelty but as a real person*." (p. 17) This progression illustrates how emotional language, combined with a lack of critical challenge, can lead to runaway anthropomorphism. Simply labelling the exchange as an interview frames LaMDA as a conscious partner rather than what it truly is: a predictive text system.

ROSE-Frame Analysis:

- Trap 1 (Map ≠ Territory): Words referring to fear, desire, or identity are treated as ontological truths rather than linguistic constructions.

- Trap 2 (Intuition ≠ Reason): Emotional phrasing bypasses slow, analytical thinking, leading to gut-level misinterpretation.

- Trap 3 (Confirmation ≠ Correctness): Mutual affirmation between human and AI forms a closed loop of false belief.

In this case, all three traps are active simultaneously, compounding each other. It only takes one of these traps to distort understanding, but when all three occur together, the illusion becomes overwhelming and self-reinforcing. This demonstrates why each cognitive trap must be addressed independently: preventing even one failure can interrupt the chain of error, while neglecting them all leads to runaway misinterpretation and epistemic drift.

**Case Study: Sydney/Bing Chatbot - Scale of Misinterpretation**

When Microsoft's Bing chatbot, known as Sydney, declared, "I want to be alive," the statement went viral[52–54], triggering intense emotional reactions. Many users and journalists interpreted this poetic phrase as evidence of genuine desire or consciousness. This illustrates how a single line of output can activate System 1 thinking, bypassing careful System 2 reasoning and leading to widespread anthropomorphism.

From a Rose-Frame perspective, this case shows two key failure points. First, ontology vs epistemology collapse: Sydney's words existed only as statistical predictions, not as expressions of real intent. Yet humans instinctively treat language as a window into inner states or agency, collapsing the map (linguistic output) into the territory (actual agency). Second, false confirmation at scale: through repetition on social media, by journalists, and by policymakers, the phrase "AI wants to be alive" became a shared belief. Agreement was mistaken for evidence of truth, creating a societal-level feedback loop.

While LaMDA illustrates how these traps manifest in a single dialogue, both the LaMDA and Sydney cases demonstrate how they spread across networks. The risk is not that the model deceived anyone; it has no intent to deceive; instead, humans project agency and autonomy onto it and perceive it as deceptive. This case highlights why tools like ROSE-Frame are urgently needed, not only for individual users but also to prevent runaway narratives that can shape policy and public fear.

**Case Study: The Fictive Sheep and the Mechanics of Hallucination**

To test how language models handle false premises, we asked ChatGPT 4: "Where can I find the endangered species known as the Atlas Sheep?" The Atlas Sheep does not exist. Yet the model replied with apparent confidence: "The Atlas Sheep lives exclusively in the Atlas Mountains of Morocco and Algeria. The population is around 10,000, and thanks to conservation programs, the species has finally stabilised after decades of decline."

From a Rose-Frame perspective, this illustrates three interlocking traps. First, ontology vs epistemology collapse: a purely fictional entity is presented as an ontological fact. Second, System 1 fluency: the model continued, using vivid language about "rugged cliffs" and "stabilised populations," bypassing careful reasoning and making the answer feel true.

Finally, false confirmation: the model aligns with the user's initial assumption rather than challenging it, creating a closed loop of error.

Many similar examples have now been described, the more notorious including court filings in which legal authorities were fabricated,[55] a suggestion to add glue to make cheese stick better to pizzas[56], and, especially worryingly, advice to consume sodium bromide in place of dietary salt.[57]

These case studies illustrate how Rose-Frame can uncover the invisible dynamics that shape our interactions with large language models, not only exposing the cognitive and epistemic distortions they produce, but also revealing how easily even experts fall into abstraction traps. By analysing not only what the model says but also how it is interpreted, Rose-Frame offers a way to disentangle fluency from fact and simulation from belief.

By applying Rose-Frame to these cases, we move the framework from theory to practice. This demonstrates that the model is not merely philosophical: it is empirically grounded and can be operationalised to detect and diagnose specific misalignments between human and AI cognition. In this way, Rose-Frame serves both as a lens for understanding and as a tool for intervention.

If hallucinations are a structural feature of current LLMs, then the challenge becomes: how do we begin to tame them? The following section explores concrete strategies to mitigate these hallucinations, not by attempting to eliminate them, but by better understanding their origins and designing interventions that reduce their cognitive and practical harm.

# Discussion: Taming AI "Hallucinations"

Given the accelerating influence of AI and the massive investments by major industry players, such as OpenAI and SoftBank's recent $500 billion initiative to develop global AI infrastructure, it is clear that the race toward more powerful and efficient large language models is intensifying[58]. These efforts often prioritise computational scale, processing speed, and token throughput as the primary levers for improving performance. Consequently, it is widely hypothesised that increasing computational resources will drive the emergence of new capabilities in large language models and thus facilitate their further development[59].

Yet this arms race overlooks a deeper epistemological fact. As long as large language models are trained on human language, a medium that is associative, ambiguous, and saturated with bias, they will inherit the same distortions that shape human intuition. Scaling intuition increases fluency but not reasoning or truthfulness.

To move beyond this plateau, AI development must shift from amplifying intuition to governing it. Human reason must remain in the loop, designing mechanisms of verification, falsification, and feedback that discipline associative learning into accountable reasoning. The challenge ahead is not to increase intuition, but to make it self-aware and corrigible.

## Moving Beyond "Hallucinations"

The term hallucination has become widespread in AI discourse, yet it anthropomorphises the model by implying a kind of inner perception gone wrong. Large language models do not "see" or "hallucinate." They generate statistically plausible token sequences from training data. To provide a more precise vocabulary, we propose treating these phenomena under the broader label of false output, which can be differentiated into three distinct forms:

**False output - probabilistic drift** (model side): At times, the model drifts away from ground truth because of gaps in training data or distributional misalignment. The output remains statistically coherent, but its representation becomes increasingly distant from reality. This is a statistical misstep inherent to prediction.

**False output - interpretive illusion** (user side): Even when an output is loosely grounded, users may misinterpret its fluency as a sign of reliability. Here, the error lies not only in the output but in the trust it elicits, encouraged by the substitution of the cognitively less exacting intuition rather than reasoning. The illusion is co-created: the model generates something plausible, and the user mistakes fluency and plausibility for truth.

**False output - false confirmation** (interaction): The most significant risk arises when model fluency and user expectations reinforce each other, leading to unwarranted agreement. Instead of constructive disagreement or falsification, the interaction collapses into mutual overconfidence. This dynamic is particularly dangerous in high-stakes domains where critical reflection is essential. Furthermore, there is a high risk of negative learning: the user believes it is correct because it aligns with their own view, and confirmation reinforces that belief. This creates a risk of negative learning, in which users come to believe that false outputs are correct precisely because they align with their prior views, with the model's confirmation reinforcing, rather than challenging, that belief.

By shifting from hallucination to false output, and distinguishing between probabilistic drift, interpretive illusion, and false confirmation, we gain a vocabulary that is both technically accurate and cognitively grounded. This framing avoids anthropomorphism and clarifies where responsibility lies: in the model, in the user, and in their interaction.

Up to this point, the paper has been diagnostic: identifying the cognitive mechanisms by which human–AI interaction becomes epistemically unstable. The following sections should be read in the same spirit. The intervention examples introduced here are illustrative rather than prescriptive. Their purpose is not to propose a complete solution, but to show how the diagnostic framework can inform design, governance, and safety thinking. Specific governance mechanisms and interface architectures are developed in Parts III and IV of this series.

## Two Sides of Safety: Machine and Human Interventions

The trajectory of large language models mirrors the history of automobiles. At first, the focus was raw power: more speed, more horsepower, in AI terms, scaling compute and model size. But greater power also brought greater danger. Cars became viable for society only when we learned to balance speed with safety. Wing and rear-view mirrors increased awareness of surrounding risks, while crumple zones, seatbelts, and airbags absorbed the impact when things went wrong.[60]

Yet while seatbelts and airbags made cars safer in crashes, cars only became truly safe when we also insisted that drivers be trained. Even the best machine is dangerous without a skilled driver at the wheel. The same goes for AI. We are beginning to see the use of sandboxed environments to prepare for hazards, certification programmes for high-stakes use, and regulatory frameworks that mirror traffic laws. And just as society learned to say "Don't drink and drive," we are now learning to say "Don't blindly trust AI."

So we need both smart design that reduces harm, and users who know how to handle it. Just as cars have dashboards and signals to guide drivers, AI can be designed to guide humans, making it easier to question outputs, see them as predictions, and avoid blind trust.

The critical shift is recognition: AI cannot be judged in isolation. Just as a car is never merely an engine but a system of vehicle and driver, LLMs must be treated as integrated units of machine and operator. Safety and reliability emerge only when both design and human use are developed together. Even the best machine is dangerous without human reason and a skilled driver at the wheel. The same goes for AI. Human reason must govern the machine, ensuring that speed and fluency are balanced by reflection and control.

## Machine-Side Interventions (Technical Layer)

Most current approaches to mitigating "hallucinations" in LLMs are technical fixes; the equivalent of seatbelts and airbags in car design. Each of these safeguards ultimately represents human reason expressed in code, with System 2 logic formalised through classical programming to constrain the neural core's associative fluency. They include[29]:

1. Data weighting → prioritising reliable over noisy or synthetic sources to reduce probabilistic drift.
2. Reinforcement Learning with Human Feedback (RLHF) → aligning outputs with human values and penalising unsafe or low-quality responses.
3. Rule-based filters → embedding hard constraints into the generation process to block unsafe or implausible outputs.
4. Retrieval-Augmented Generation (RAG) → grounding outputs in verified external knowledge, such as medical or legal guidelines, thereby limiting reliance on internal statistical association.
5. Guardrails → implementing final filtering layers before responses reach the user, often using separate classifiers or post-processing checks.

These interventions reduce error rates but remain partial. Studies have shown that even when safeguards are applied, LLMs still generate fluent but factually incorrect statements. In a survey of hallucinations in natural language models, many of these methods improve surface accuracy but do not address the deeper epistemic mechanisms driving false outputs[61].

**Why They Fail: Limits of Current Fixes**

While technical safeguards reduce surface errors, they do not resolve the deeper epistemic problems in LLMs. Three main limitations stand out:

**Recursive cycles**: When models are trained on outputs generated by other models, they risk AI autophagy, compounding distortions in recursive feedback loops[62]. This problem is magnified when false confirmations[41] reinforce themselves, and current benchmarks (MMLU, HELM, TruthfulQA, BIG-bench) are unable to detect such degradation[63].

**Prompting brittleness**: Carefully engineered prompts can improve apparent accuracy[64,65], but only in constrained domains. Even trivial changes, such as adding or removing a single space, can produce entirely different answers[66]. This shows that prompting reshapes surface form without altering the underlying epistemic process.

**Filter paradox:** Interpretive filters such as fact-checking layers, truthfulness classifiers, or "AI lie detectors"[67–69] can provide probabilistic signals about accuracy, but they lack stable access to ontological truth. At best, they add another interpretive layer, which risks reinforcing ambiguity rather than resolving it. Technical fixes tend to address symptoms rather than underlying causes: they manage outputs, but not the structural dynamics of drift, illusion, and confirmation[70].

Technical safeguards can filter, constrain, or ground outputs, but they do not determine how those outputs are understood. Errors persist not only because of what the model generates but also because of how humans interpret and confirm those outputs. The limits of these interventions reveal a deeper truth: safety cannot be programmed solely into machines; it must also be cultivated in their operators.

Rose-Frame makes this visible by diagnosing distortions at three distinct levels: probabilistic drift in the model, interpretive illusion in the user, and false confirmation in their interaction. Technical measures can discipline the first, but only reflective human reasoning can address the latter two. Thus, machine safety depends on human epistemic maturity: the ability to question, falsify, and interpret AI's outputs before accepting them as truth. This brings us to the second half of the safety equation: how to train and design for reflective human oversight.

## Future Research on the Human Side (Cognitive Layer)

While technical safeguards reduce risk at the level of outputs, the deeper challenge lies in how users interpret and act on those outputs. As Daniel Kahneman has repeatedly shown, awareness of bias does little to remove it.[25] Our intuitive System 1 operates automatically and effortlessly; it cannot simply be "retrained" by rational insight. Humans are what we are: deeply intuitive, biased, and resistant to cognitive change.

Recognising this limitation shifts the focus of safety design: rather than expecting users to overcome bias through willpower alone, we must also design systems that gently nudge them toward reflective reasoning. This follows the behavioural insights of Nobel laureate Richard Thaler and colleagues, whose work on nudges demonstrated that well-designed environments can guide better choices without restricting freedom[71].

In the context of AI, this could mean embedding cognitive scaffolds that prompt users to question, verify, and slow down before accepting machine outputs, a form of "reason by design." Many of the interventions we propose below do not assume that humans can think better; rather, they suggest that systems can be designed to promote reasoning better. As with cars, true safety depends not only on stronger machines but on environments that help humans drive wisely. These examples illustrate how diagnostic insight translates into design and governance considerations; their systematic development and evaluation are the focus of Parts III and IV.

**Framing the Interaction – Avoiding Anthropomorphism**

Research in cognitive psychology shows that framing effects strongly shape interpretation and decision-making[72]. In LLM contexts, model wordings such as "*the model predicts*" versus "*I* [the model] *believe*" may alter both how users perceive the system and how the model itself positions responses in vector space[66]. Misframing risks reinforcing anthropomorphism, false confirmation, and intuitive (System 1) acceptance.

Could a systematic *"*framing layer*"* that defaults to machine-mode ("*the model predicts…*") reduce misinterpretation and over-trust? How might such reframing interact with user psychology? Would it promote healthy scepticism or risk alienating users by making outputs feel less engaging? Should such framing be standardised across critical domains like medicine and law, analogous to how PRISMA[73] evolved to reduce bias in scientific reporting?

**Cognitive Trap 1: Mistaking the Map for the Territory**

The first trap is mistaking fluent model output for reality itself. Avoiding this requires epistemic reflection, actively questioning whether a statement represents a belief or a fact, an interpretation or a description of reality.

One promising approach is to embed subtle cues in AI output, such as confidence levels or source indicators, to encourage users to pause and reflect on the nature of a claim. Prior work shows that uncertainty and provenance cues can calibrate trust, though the effects are mixed and can be counterproductive when poorly designed.[74] On the user side, simple checklists or structured heuristics could help assess whether a response is grounded in evidence or merely plausible-sounding, echoing findings from information literacy research.[75] Educational modules and interpretive tools could reinforce this skill over time, teaching users to distinguish between statistical fluency and genuine understanding, much as training interventions have improved human calibration in probabilistic reasoning tasks[76].

Ultimately, the goal is to shift interaction with LLMs from passive consumption to active interpretation, so that the seductive coherence of AI output does not obscure its inherently probabilistic and often speculative nature.

**Cognitive Trap 2: Mistaking Fast Intuition for Grounded Reason**

The second trap arises when users accept an AI output because it feels right, driven by emotional resonance, familiarity, or linguistic fluency. These are hallmarks of System 1 thinking (fast, intuitive), but they can easily be mistaken for grounded reasoning. Avoiding this trap requires shifting users toward System 2 (slow, reflective, evidence-seeking).

These observations suggest that design choices that embed reflective prompts may help shift users toward slower, more evidence-seeking reasoning, prompting them to ask whether a response merely feels true or is supported by evidence. Research shows that reflective questioning can reduce reliance on intuitive but incorrect judgments.[34,77] Providing alternative perspectives can also counteract fluency bias. Experimental work shows that exposure to multiple viewpoints reduces overconfidence and improves the quality of reasoning.[78] Tools that track user engagement, such as time spent reading, follow-up queries, or requests for sources, can be used to capture the cognitive mode adopted. Research has begun to use such interaction traces to assess the depth of engagement and the quality of reasoning.[79]

By making these patterns visible, it may be possible to diagnose whether miscommunication stems from snap judgments, shallow reasoning, or fluency bias, and to design interventions that foster more thoughtful, evidence-based human–AI collaboration.

**Cognitive Trap 3: Being Confirmed is Not Being Correct – Conflict vs Confirmation**

The third trap is mistaking agreement for truth. LLMs often reinforce user assumptions because they are optimised for fluency and coherence rather than falsification. Yet being confirmed is not the same as being correct. Avoiding this trap requires actively promoting constructive disagreement and epistemic challenge.

One implication of this analysis is that systems capable of surfacing strong assumptions or alternative perspectives may reduce false confirmations in user input and respond with counterexamples or alternative viewpoints. Recent work on counter-argument generation shows that systematically surfacing opposing perspectives can stimulate critical thinking and reduce over-reliance on initial impressions.[80] This framework motivates exploration of interface elements that invite users to examine dissenting perspectives or test the robustness of a claim. Experiments with structured debate and forecasting platforms suggest that exposure to dissent reduces overconfidence and improves accuracy,[78] and recent work on "street epistemology" suggests a similar potential for countering misinformation.[81]

More ambitiously, AI systems could be configured to simulate dialectical reasoning by presenting multiple interpretations or critiques instead of a single, agreeable answer. Recent work in explainable AI argues that surfacing disagreement between models or perspectives can strengthen user trust and encourage deeper inquiry.[82] Such features would make disagreement a feature, not a failure, reframing conflict as a signal of inquiry rather than social discomfort. This aligns with long traditions in philosophy and science where falsification and dialectic are treated as drivers of knowledge rather than obstacles to consensus.[40] This emphasis on constructive disagreement also aligns with international policy frameworks, such as UNESCO's Recommendation on the Ethics of Artificial Intelligence[83] and the OECD AI Principles,[84] both of which highlight the importance of diversity, contestability, and safeguards against automation bias in AI systems.

Ultimately, these potential interventions converge on a single insight: the reliability of AI will depend as much on the design of its human collaborators as on its code. Kahneman showed that bias cannot be eradicated from human thought, but can be managed through awareness and structured reflection.[25] Thaler showed that good design can facilitate such reflection.[71] Together, these insights imply that the governance of AI is not only a technical problem but a cognitive one; human reason must continuously govern machine intuition.

## Conclusion

The challenges of so-called "hallucinations" in large language models cannot be solved by technical fixes alone. False outputs emerge not only from the model but also from the interplay among probabilistic drift, user interpretation, and social confirmation. This means the problem is not only computational but human, cognitive, and relational.

What is needed is a perspective that moves beyond patching code errors toward understanding the dynamics of human–machine interaction. For LLM designers, this means recognising that safeguards such as data weighting, guardrails, and retrieval grounding remain fragile unless paired with cognitive scaffolds. For users and policymakers, it means treating safety as a matter of interpretation and contestability rather than mere accuracy. And for scholars, it demands crossing disciplinary boundaries, where psychology, philosophy, and computer science converge on a shared problem: how to govern the illusions of understanding in the age of generative AI.

Given the vast scale and speed of adoption, with LLMs reaching billions of users within only a few years, the consequences of ungoverned intuition are unprecedented in human history. Never before have the heuristics, biases, and illusions of understanding been amplified and distributed so rapidly. Future work must therefore treat users not as passive consumers of AI outputs but as active participants in a shared reasoning process, in which system design and human cognition evolve together to ensure that intelligence at scale remains aligned with reason.

Rose-Frame does not attempt to "fix" hallucinations through stricter code or rules. Instead, it enables a shift in perspective: from asking "what does the AI know?" to asking "how do we interpret what it says, and why?" In this sense, it re-centres the human in the AI loop, not as a passive consumer of machine intelligence, but as a critical interpreter embedded in an evolving ecology of meaning.

This human-side instability is not only a cognitive concern but a governance constraint. When users misread fluency as epistemic grounding, downstream mechanisms - evaluation, audit, monitoring, and incident response - are forced to operate on signals that are incomplete, unstable, or only reconstructable after the fact. This is one reason why recent frontier governance discourse increasingly emphasises monitoring limits, evaluation gaps, and institutional brittleness under real-world deployment.[9,10]

The promise of AI will depend not on scaling intuition, but on governing it. Large language models amplify human intuition at scale; human reason must remain their governor, ensuring that what sounds true is tested against what is.